\documentclass{article}
\usepackage{times}
\usepackage{hyperref,color}
\definecolor{mydarkblue}{rgb}{0,0.08,0.45}
\hypersetup{
  colorlinks=true,
  linkcolor=mydarkblue,                                                          
  citecolor=mydarkblue,                                                          
  filecolor=mydarkblue,                                                          
  urlcolor=mydarkblue,
}

\usepackage{natbib}
\usepackage{url}
\usepackage{import, subfiles}
\usepackage{tabularx}
\usepackage{algorithm}
\usepackage{algorithmic}
\usepackage{multirow}
\usepackage{fullpage}
\usepackage{latex-defs}
\usepackage{enumitem}
\usepackage{bbm}
\usepackage{arxiv} 

\newcommand{\bi}{\mathbbm{1}}
\newcommand{\yold}{y_{t-1}}
\newcommand{\zold}{z_{t-1}}
\newcommand{\ynew}{y_t}
\newcommand{\znew}{z_t}
\newcommand{\apx}{the appendix}

\newcommand\theTitle{Learning Fast-Mixing Models for Structured Prediction}

\icmltitlerunning{\theTitle}

\begin{document}

\twocolumn[
  \icmltitle{\theTitle}

\icmlauthor{Jacob Steinhardt}{jsteinhardt@cs.stanford.edu}
\icmlauthor{Percy Liang}{pliang@cs.stanford.edu}
\icmladdress{Stanford University,
             353 Serra Street, Stanford, CA 94305 USA}

\vskip 0.3in
]

\begin{abstract}
Markov Chain Monte Carlo (MCMC) algorithms are often used for approximate
inference inside learning, but their slow mixing can be difficult to diagnose
and the approximations can seriously degrade learning.
To alleviate these issues, we define a new model family
using strong Doeblin Markov chains,
whose mixing times can be precisely controlled by a parameter.
We also develop an algorithm to learn such models,
which involves maximizing the data likelihood under the induced stationary
distribution of these chains.
We show empirical improvements on two challenging inference tasks.
\end{abstract}

\section{Introduction}
\label{sec:intro}

Conventional wisdom suggests that rich features and
highly-dependent variables necessitate intractable inference.
Indeed, the dominant paradigm is to first
define a joint model, and then use approximate inference (e.g., MCMC) to learn
that model.
While this recipe can generate good results in practice,
it has two notable drawbacks:
(i) diagnosing convergence of Markov chains is extremely difficult
\citep{gelman1992single,cowles1996markov};
and (ii) approximate inference can be highly suboptimal
in the context of learning \citep{wainwright2006estimating,kulesza2007structured}.

In this paper, we instead use MCMC to \emph{define} the model family itself:
For a given $T$, we construct a family of Markov
chains using arbitrary rich features,
but whose mixing time is \emph{guaranteed} to be at most $O(T)$.
The corresponding stationary distributions make up the model family.
We can think of our Markov chains as parameterizing a family of
``computationally accessible'' distributions,
where the amount of computation is controlled by $T$.

\

For concreteness, suppose we are performing a structured prediction task
from input $x$ to a complex output $y$.
We construct Markov chains of the following form, called \emph{strong Doeblin
chains} \citep{doeblin1940elements}:
\begin{align}
  \label{eqn:construction}
\tilde{A}_{\theta}(\ynew \mid \yold, x) = (1-\epsilon)A_{\theta}(\ynew \mid \yold, x) + \epsilon \, u_\theta(\ynew \mid x),
\end{align}
where $\epsilon$ is a mixture coefficient and $\theta$ parameterizes 
$A_\theta$ and $u_\theta$.
Importantly, $u_\theta$ does not depend on the previous state $\yold$.
For intuition, think of $u_\theta$ as a simple tractable model
and $A_\theta$ as Gibbs sampling in a complex intractable model.
With probability $1-\epsilon$, we progress according to
$A_{\theta}$, and with probability $\epsilon$ we draw a fresh sample from 
$u_{\theta}$, which performs an informed random restart.
When $\epsilon = 1$, we are drawing i.i.d. samples from $u_{\theta}$; 
we therefore mix in a single step, but our stationary distribution must
necessarily be very simple.
When $\epsilon = 0$, the stationary distribution can be much 
richer, but we have no guarantees on the mixing time. For intermediate values 
of $\epsilon$, we trade off between representational power and mixing time.

A classic result is that a given strong Doeblin chain mixes in time at most
$\frac{1}{\epsilon}$ \citep{doeblin1940elements}, and that we can draw
an exact sample from the stationary distribution in
expected time $O(\frac{1}{\epsilon})$ \cite{corcoran1998perfect}.
In this work, we prove new results that help us understand the strong Doeblin model families.
Let $\sF$ and $\tilde\sF$ be the family of stationary distributions
corresponding to $A_{\theta}$ and $\tilde A_\theta$
as defined in (\ref{eqn:construction}), respectively.
Our first result is that as $\epsilon$ decreases,
the stationary distribution of any $\tilde{A}_{\theta}$ 
monotonically approaches the stationary distribution of
the corresponding $A_{\theta}$
(as measured by either direction of the KL divergence).
Our second result is that if $\frac{1}{\epsilon}$ is much larger
than the mixing time of $A_\theta$,
then the stationary distributions of $A_\theta$ and $\tilde A_\theta$
are close under a certain Mahalanobis distance.
This shows that any member of $\sF$
that is computationally accessible via the Markov chain
is well-approximated by its counterpart in $\tilde{\sF}$.
\begin{center}
\begin{tikzpicture}[scale=1.1]
\def\s{1in}
\def\ss{1em}
\node[draw=blue!50!black,fill=blue!50,fill opacity=0.9,circle,minimum height=0.4*\s] (F0) at (-0.05,-0.29) {};
\node[draw=blue,fill=blue,fill opacity=0.5,circle,minimum height=0.7*\s] (F) at (0,0) {};
\node at ([xshift=0.4*\ss,yshift=-0.6*\ss]F.north west) {\small{$\sF$}};
\coordinate (p) at (0.38, -0.03);
\draw[rotate around={-45:(p)},draw=red!50!black,fill=red!70!black,fill opacity=0.4] (p) ellipse (0.25in and 0.42in);
\node at ([xshift=0.235in,yshift=0.235in]p) {\small{$\tilde{\sF}$}};
\node at ([xshift=-0.6*\ss,yshift=-0.6*\ss]F0.north east) {\small{$\sF_0$}};
\end{tikzpicture}
\end{center}
The figure above shows $\sF$ and $\tilde\sF$, together with the subset $\sF_0$ of $\sF$
whose Markov chains mix quickly.
$\tilde{\sF}$ (approximately) covers $\sF_0$, and contains some distributions
outside of $\sF$ entirely.

In order to learn over $\tilde{\sF}$, we show how to maximize the 
likelihood of the data
under the stationary distribution of $\tilde A_\theta$.
Specifically,
we show that we can compute a stochastic gradient of the log-likelihood in
expected time $O(\tfrac{1}{\epsilon})$.
Thus, in a strong sense, our objective function explicitly accounts for computational
constraints.

We also generalize strong Doeblin chains,
which are a mixture of two base chains, $u_\theta$ and $A_\theta$,
to \emph{staged} strong Doeblin chains,
which allow us to combine more than two base chains.
We introduce an auxiliary variable $z$ representing the ``stage''
that the chain is in.
We then transition between stages, using the base chain corresponding to the current
stage $z$ to advance the concrete state $y$.
A common application of this generalization is defining a sequence of increasingly
more complex chains, similar in spirit to annealing.
This allows sampling to become gradually more sensitive to the structure of the problem.

We evaluated our methods on two tasks:
(i) inferring words from finger gestures on a 
touch screen and (ii) inferring DNF formulas for program verification. 
Unlike many structured prediction problems where local potentials provide
a large fraction of the signal, in the two tasks above, local potentials offer
a very weak signal; reasoning carefully about the higher-order potentials
is necessary to perform well.
On word inference, we showed that learning strong Doeblin chains
obtained a 3.6\% absolute improvement in character accuracy over Gibbs sampling 
while requiring 5x fewer samples.
On DNF formula inference, our staged strong Doeblin chain obtains
an order of magnitude speed improvement over plain Metropolis-Hastings.

To summarize, the contributions of this paper are: 
We formally define a family of MCMC algorithms based on strong Doeblin
chains with guaranteed fast mixing times
      (Section~\ref{sec:defs-examples}).
We provide an extensive analysis of the theoretical properties of this 
      family (Section~\ref{sec:theory}), together with a generalization to 
      a staged version (Section~\ref{sec:hierarchical}).
We provide an algorithm for learning the parameters of a strong Doeblin chain
      (Section~\ref{sec:learning}).
We demonstrate superior experimental results relative to baseline MCMC samplers
      on two tasks, word inference and DNF formula synthesis
      (Section~\ref{sec:experiments}).

\section{A Fast-Mixing Family of Markov Chains}
\label{sec:defs-examples}

Given a Markov chain with 
transition matrix $A(\ynew \mid \yold)$ and a distribution $u(\ynew)$, define 
a new Markov chain with transitions given by 
$\tilde{A}(\ynew \mid \yold) \eqdef (1-\epsilon)A(\ynew \mid \yold) + \epsilon u(\ynew)$. 
(We suppress the dependence on $\theta$ and $x$ for now.)
In matrix notation, we can write $\tilde{A}$ as
\begin{align}
  \label{eqn:constructionMatrix}
\tilde{A} \eqdef (1-\epsilon)A + \epsilon u \bi^{\top}.
\end{align}
In other words, with probability $\epsilon$ we restart from 
$u$; otherwise, we transition according to $A$. Intuitively, 
$\tilde{A}$ should mix quickly because a restart from $u$ renders
the past independent of the future (we formalize this in Section~\ref{sec:theory}).
We think of $u$ as a simple tractable model that 
provides \emph{coverage}, and $A$ as a complex model that provides \emph{precision}.
\subfile{fig-simple.stex}

\paragraph{Simple example.} 
To gain some intuition, we work through a simple example with the Markov chain $A$ depicted in 
Figure~\ref{fig:simple}. The stationary distribution of this chain is $\left[ \begin{array}{ccc} \frac{1}{2+3\delta} & \frac{3\delta}{2+3\delta} & \frac{1}{2+3\delta} \end{array} \right]$,
splitting most of the probability mass evenly between states $1$ and $3$. The 
mixing time of this chain is approximately $\frac{1}{\delta}$, since once the chain falls into 
either state $1$ or state $3$, it will take about $\frac{1}{\delta}$ steps for it to escape back 
out. If we run this Markov chain for $T$ steps with $T \ll \frac{1}{\delta}$, then our samples 
will be either almost all in state $1$ or almost all in state $3$, and thus will provide a poor 
summary of the distribution. If instead we perform random restarts with probability $\epsilon$ 
from a uniform distribution $u$ over $\{1,2,3\}$, then the restarts give us the opportunity to 
explore both modes of the distribution. After a restart, however, the chain will more likely fall into 
state $3$ than state $1$ ($\frac{5}{9}$ probability vs. $\frac{4}{9}$), so for 
$\epsilon \gg \delta$ the stationary distribution will be noticeably perturbed by the restarts.
If $\epsilon \ll \delta$, then there will be enough time for the chain to mix between restarts,
so this bias will vanish.
See Figure~\ref{fig:simple} for an illustration of this phenomenon.

\section{Theoretical Properties}
\label{sec:theory}
Markov chains that can be expressed according to (\ref{eqn:constructionMatrix})
are said to have 
\emph{strong Doeblin parameter} $\epsilon$ \citep{doeblin1940elements}.
In this section, we characterize the stationary distribution and mixing time of 
$\tilde{A}$, and also relate the stationary distribution of $\tilde{A}$ to that of
$A$ as a function of $\epsilon$.
Often the easiest way to study the mixing time of $\tilde{A}$ 
is via its spectral gap, which is defined as $1-\lambda_2(\tilde A)$, where 
$\lambda_2(\tilde A)$ is the second-largest eigenvalue (in complex norm). A standard 
result for Markov chains is that, under mild assumptions, the mixing time of $\tilde{A}$ is 
$O\left(\frac{1}{1-\lambda_2(\tilde A)}\right)$. 
We assume throughout this section that 
$A$ is ergodic but not necessarily that it is reversible. See 
Section 12.4 of \citet{levin2009markov} for more details.

Our first result relates the spectral gap (and hence the mixing time) to 
$\epsilon$. This result (as well as the next) are well-known but we include them 
for completeness. For most results in this section, we sketch the proof here and 
provide the full details in \apx.
\begin{proposition}
\label{prop:mixing}
The spectral gap of $\tilde{A}$ is at least $\epsilon$; that is, $1-\lambda_2(\tilde A) \ge \epsilon$. In particular, $\tilde{A}$ mixes in time $O(\frac1\epsilon)$.
\end{proposition}
The key idea is that all eigenvectors of $\tilde{A}$ and $A$ (except for the 
stationary distribution) are equal, and that 
$\lambda_k(\tilde{A}) = (1-\epsilon)\lambda_k(A)$ for $k>1$. 

Having established that $\tilde{A}$ mixes quickly, the next step is to 
determine its stationary distribution:
\begin{proposition}
\label{prop:stationary}
Let $\tilde{\pi}$ be the stationary distribution of $\tilde{A}$. Then 
\[ \tilde{\pi} = \epsilon \sum_{j=0}^{\infty} (1-\epsilon)^j A^ju = \epsilon(I-(1-\epsilon)A)^{-1}u. \]
\end{proposition}
This can be directly verified algebraically.
The summation over $j$ shows that
we can in fact draw an exact sample from $\tilde{\pi}$ by drawing 
$T \sim \Geometric(\epsilon)$,\footnote{If $T \sim \Geometric(\epsilon)$, we have $\mathbb P[T = j] = \epsilon(1-\epsilon)^j$ for $j\geq 0$.}
initializing from $u$, and transitioning $T$ times according to $A$.
This is intuitive, since at a generic point in time we expect the most 
recent sample from $u$ to have occurred $\Geometric(\epsilon)$ steps ago.
Note that $\mathbb E[T+1] = \frac1\epsilon$,
which is consistent with
the fact that the mixing time is $O(\frac1\epsilon)$ (Proposition~\ref{prop:mixing}). 

We would like to relate the stationary distributions $\tilde{\pi}$ and 
$\pi$ of $\tilde{A}$ and $A$. The next two results (which are new) do so. 

Let $\tilde{\pi}_{\epsilon}$ denote the stationary distribution of 
$\tilde{A}$ at a particular value of $\epsilon$; note that $\tilde{\pi}_1 = u$ 
and $\tilde{\pi}_0 = \pi$. We will show that $\tilde{\pi}_{\epsilon}$ approaches 
$\pi$ \emph{monotonically}, for both 
directions of the KL divergence. In particular, for any $\epsilon < 1$, 
$\tilde{\pi}_{\epsilon}$ is at least as good as $u$ at approximating $\pi$.

To show this, we make use of the following 
lemma from \citet{murray2008notes}:

\begin{lemma}
\label{lem:klmc}
If $B$ is a transition matrix with stationary distribution $\pi$, then 
$\KL{\pi}{B\pi'} \leq \KL{\pi}{\pi'}$ and $\KL{B\pi'}{\pi} \leq \KL{\pi'}{\pi}$.
\end{lemma}
Using this lemma, we can prove the following monotonicity result:
\begin{proposition}
\label{prop:monotonic}
Both $\KL{\tilde{\pi}_{\epsilon}}{\pi}$ and $\KL{\pi}{\tilde{\pi}_{\epsilon}}$ 
are monotonic functions of $\epsilon$.
\end{proposition}
The idea is to construct a transition matrix $B$ that maps 
$\tilde{\pi}_{\epsilon_1}$ to 
$\tilde{\pi}_{\epsilon_2}$ for given $\epsilon_2 < \epsilon_1$, then show that 
its stationary distribution is $\pi$ and apply Lemma~\ref{lem:klmc}.

With Proposition~\ref{prop:monotonic} in hand, a natural next question is how 
small $\epsilon$ must be before $\tilde{\pi}$ is reasonably close to $\pi$. 
Proposition~\ref{prop:mahalanobis} provides one such bound: 
$\tilde{\pi}$ is close to $\pi$ if $\epsilon$ is small compared to the spectral gap 
$1-\lambda_2(A)$. 
\begin{proposition}
\label{prop:mahalanobis}
Suppose that $A$ satisfies detailed balance with respect to $\pi$. 
Let $\tilde{\pi}$ be the stationary distribution of 
$\tilde{A}$.
Define $d_{\pi}(\pi') \eqdef \|\pi-\pi'\|_{\diag(\pi)^{-1}} = \sqrt{-1 + \sum_{y} \frac{\pi'(y)^2}{\pi(y)}}$, 
where $\|v\|_{M}$ is the Mahalonobis distance $\sqrt{v^{\top}Mv}$.
Then $d_{\pi}(\tilde{\pi}) \leq \frac{\epsilon}{1-\lambda_2(A)} \cdot d_{\pi}(u)$. 
(In particular, $d_{\pi}(\tilde{\pi}) \ll 1$ if $\epsilon \ll (1-\lambda_2(A))/d_{\pi}(u)$.)
\end{proposition}
The proof is somewhat involved, but the key step is to establish that 
$d_\pi(\pi')$ is convex in $\pi'$ and contractive with respect to $A$ 
(more precisely, that $d_\pi(A\pi') \leq \lambda_2(A)d_{\pi}(\pi')$).

Proposition~\ref{prop:mahalanobis} says that if $A$ mixes quickly, then $\tilde{A}$ 
and $A$ will have similar stationary distributions. This serves as a sanity check: 
if $A$ already mixes quickly, then $\tilde{\pi}$ is a good approximation to $\pi$; otherwise, 
the Doeblin construction ensures that we are at least converging to \emph{some} distribution, 
which by Proposition~\ref{prop:monotonic} approximates $\pi$ at least as well as $u$ does.

\subsection{Staged strong Doeblin chains}
\label{sec:hierarchical}
\subfile{fig-hierarchical.stex}

Recall that to run a strong Doeblin chain $\tilde A$,
we first sample from $u$, and then transition according to 
$A$ for approximately $\frac{1}{\epsilon}$ steps.
The intuition is that sampling 
from the crude distribution $u$ faciliates global exploration of the state space,
while the refined transition $A$ hones in on a mode.
However, for complex problems,
there might be a considerable gap between what is possible with exact inference ($u$) and
what is needed for accurate modeling ($A$).
This motivates using multiple stages of MCMC to bridge the gap.

To do this, we introduce an auxiliary variable $z \in \sZ$ denoting which stage of 
MCMC we are currently in.  For each stage $z$, we have a Markov chain $A_z(\ynew \mid
\yold)$ over the original state space.  We also define a Markov chain $C(\znew \mid \zold)$
over the stages.  To transition from $(\yold,\zold)$ to $(\ynew,\znew)$, we first
sample $\znew$ from $C(\znew \mid \zold)$ and then $\ynew$ from $A_{\znew}(\ynew \mid \yold)$.
If there is a special state $z^*$ for which $A_{z^*}(\ynew \mid \yold) = u(\ynew)$ (i.e., $A_{z^*}$ does not depend on 
$\yold$), then we call the resulting chain a \emph{staged strong Doeblin chain}.

For example, if $z \in \{0,1\}$
and we transition from $0$ to $1$ with probability $1-\epsilon$ and from $1$ to 
$0$ with probability $\epsilon$, then we recover strong Doeblin chains assuming $z^* = 0$
(Figure~\ref{fig:hierarchical}(a)). 
As another example (Figure~\ref{fig:hierarchical}(b)), let $z \in \{0,1,2\}$.
When $z = z^* = 0$, transition according to a restart 
distribution $u \bi^{\top}$; when $z = 1$, transition according to a simple chain $A_1$;
and when $z = 2$ transition according to a more complex chain $A_2$.
If we transition from $z=0$ to $z=1$ with 
probability $1$, from $z=1$ to $z=2$ with probability $\epsilon_1$, 
and from $z=2$ to $z=0$ with probability $\epsilon_2$, then we will on average 
draw $1$ sample from $u$, $\frac{1}{\epsilon_1}$ samples from $A_1$, and 
$\frac{1}{\epsilon_2}$ samples from $A_2$.

We now show that staged strong Doeblin chains mix quickly as long as 
we visit $z^*$ reasonably often.
In particular, the following theorem 
provides guarantees on the mixing time
that depend only on $z^*$ and on the structure of $C(\znew \mid \zold)$, analogous 
to the previous dependence only on $\epsilon$.
The condition of the theorem asks for times $a$ and $b$ such that the first time 
after $a$ that we hit $z^*$ is almost independent of the starting state $z_0$, 
and is less than $b$ with high probability.
\begin{theorem}
\label{thm:hierarchical}
Let $M$ be a staged strong Doeblin chain on $\sZ \times \sY$.
Let $\tau_a$ be the earliest time $s \geq a$ for which $z_s = z^*$. Let 
$\beta_{a,s} = \min_{z \in \sZ} \bP[\tau_a = s \mid z_0 = z]$ and
$\gamma_{a,b} \eqdef \sum_{s=a}^b \beta_{a,s}$.
Then $M^b$ has strong Doeblin parameter $\gamma_{a,b}$. In particular, 
the spectral gap of $M$ is at least $\frac{\gamma_{a,b}}{b}$. (Setting 
$a = b = 1$ recovers Proposition~\ref{prop:mixing}.)
\end{theorem}
The key idea is that, conditioned on $\tau_a$, 
$(y_b,z_b)$ is independent of $(y_0,z_0)$ for all $b \geq \tau_a$.
For the special case that the stages form a cycle as in Figure~\ref{fig:hierarchical}, 
we have the following corollary:
\begin{corollary}
\label{cor:hierarchical}
Let $C$ be a transition matrix on $\{0,\ldots,k-1\}$ such that 
$C(\znew = i \mid \zold = i) = 1-\delta_i$ and $C(\znew = (i+1) \text{ \rm mod } k \mid \zold = i) = \delta_i$. 
Suppose that $\delta_{k-1} \leq \frac{1}{\max(2,k-1)}\min\{\delta_0,\ldots,\delta_{k-2}\}$. 
Then the spectral gap of the joint Markov chain is at least 
$\frac{\delta_{k-1}}{78}$.
\end{corollary}
The key idea is that, when restricting to the time interval 
$[2/\delta_{k-1},3/\delta_{k-1}]$, the time of first transition from $k-1$ 
to $0$ is approximately $\Geometric(\delta_{k-1})$-distributed (independent 
of the initial state), which allows us 
to invoke Theorem~\ref{thm:hierarchical}.
We expect the optimal constant to be much smaller than $78$.

\section{Learning strong Doeblin chains}
\label{sec:learning}
Section~\ref{sec:theory} covered properties of strong Doeblin chains
$(1-\epsilon) A_\theta + \epsilon u_\theta \bi^\top$ for a fixed parameter vector $\theta$.
Now we turn to the problem of learning $\theta$ from data.
We will focus on the discriminative learning setting
where we are given a dataset $\{(x^{(i)}, y^{(i)})\}_{i=1}^n$ 
and want to maximize the conditional log-likelihood:
\[ \mathcal O(\theta) = \frac{1}{n} \sum_{i=1}^n \log p_{\theta}(y^{(i)} \mid x^{(i)}), \]
where now $p_{\theta}$ is the stationary distribution of 
$\tilde{A}_{\theta} = (1-\epsilon)A_{\theta} + \epsilon u_{\theta} \bi^{\top}$.
We assume for simplicity that the chains $A_\theta$ and $u_\theta$ are given by 
conditional exponential families:
\begin{align}
\notag
A_{\theta}(y \mid y', x) &\eqdef \exp\left(\theta^{\top} \phi(x, y', y) - \log Z(\theta; x, y)\right), \\ 
u_{\theta}(y \mid x) &\eqdef \exp\left(\theta^{\top} \phi(x, y) - \log Z(\theta; x)\right), 
\end{align}
where each $\phi$ outputs a feature vector and the $Z$ are partition functions.
By Proposition~\ref{prop:mixing}, $\tilde{A}_{\theta}$ mixes quickly 
for all $\theta$. On the other hand, the 
parameterization of $A_{\theta}$ captures a rich family of transition kernels, 
including Gibbs sampling.

At a high level, our learning algorithm performs stochastic gradient descent
on the negative log-likelihood.  However, the negative log-likelihood is only 
defined implicitly in terms of the stationary distribution of a Markov chain, 
so the main challenge is to show that it can be computed 
efficiently.
To start, we assume that we can
operate on the base chains $u_\theta$ and $A_\theta$ for one step efficiently:
\begin{assumption}
\label{assumption:minimal}
We can efficiently 
sample $y$ from $u_{\theta}(\cdot \mid x)$ and 
$A_{\theta}(\cdot \mid y', x)$, as well as compute
$\frac{\partial \log u_{\theta}(y \mid x)}{\partial \theta}$ and 
$\frac{\partial \log A_{\theta}(y \mid y', x)}{\partial \theta}$.
\end{assumption}
Under Asssumption~\ref{assumption:minimal}, we will show how to efficiently 
compute the gradient of 
$\log p_{\theta}(y^{(i)} \mid x^{(i)})$ with respect to $\theta$. 
The impatient reader may skip ahead to the final pseudocode, 
which is given in 
Algorithm~\ref{alg:compute-gradient}.

For convenience, we will suppress the dependence on $x$ and $i$ and 
just refer to $p_{\theta}(y)$ instead of $p_{\theta}(y^{(i)} \mid x^{(i)})$. 
Computing the gradient  of $\log p_{\theta}(y)$
is non-trivial, since the formula for $p_{\theta}$ is somewhat involved:
\[ p_{\theta}(y) = \epsilon \sum_{j=0}^{\infty} (1-\epsilon)^j [A_{\theta}^ju_{\theta}](y). \]
We are helped by the following generic identity on gradients of conditional 
log-probabilities, proved in \apx.
\begin{lemma}
\label{lem:gradient}
Let $z$ have distribution $p_{\theta}(z)$ parameterized by a vector $\theta$. 
Let $S$ be any measurable set. Then
\[ \frac{\partial \log p_{\theta}(z \in S)}{\partial \theta} = \bE_{z}\left[ \frac{\partial \log p_{\theta}(z)}{\partial \theta} \middle| z \in S\right]. \]
\end{lemma}

We can utilize Lemma~\ref{lem:gradient} by interpreting $y \mid \theta$ as the 
output of the following generative process, which by Proposition~\ref{prop:stationary} 
yields the stationary distribution of $\tilde{A}_{\theta}$:
\begin{itemize}[itemsep=2pt,topsep=0pt,parsep=0pt,partopsep=0pt,leftmargin=18pt]
\item Sample $y_0$ from $u_{\theta}$ and $y_{t+1} \mid y_t$ from 
      $A_{\theta}$ for $t=0,1,\ldots$.
\item Sample $T \sim \Geometric(\epsilon)$ and let $y = y_T$. 
\end{itemize}
We then invoke Lemma~\ref{lem:gradient} with $z = (T,y_{0:T})$ and $S$ encoding the event that 
$y_T = y$. As long as we can sample from the posterior distribution of $(T, y_{0:T})$ 
conditioned on $y_T = y$, we can 
compute an estimate of $\frac{\partial}{\partial \theta} \log p_{\theta}(y)$ 
as follows:
\begin{itemize}[itemsep=2pt,topsep=0pt,parsep=0pt,partopsep=0pt,leftmargin=18pt]
\item Sample $(T, y_{0:T}) \mid y_T = y$.
\item Return $\frac{\partial \log p_{\theta}(T, y_{0:T})}{\partial \theta} = \frac{\partial \log u_{\theta}(y_0)}{\partial \theta}$\\ \phantom{xxxxxxxxxxxxxxxxxxx}  $+ \sum_{t=1}^T \frac{\partial \log A_{\theta}(y_t \mid y_{t-1})}{\partial \theta}$.
\end{itemize}
\subsection{Sampling schemes for $(T, y_{0:T})$}
\label{sec:sampling}
By the preceding comments, it suffices to construct a sampler
for $(T, y_{0:T}) \mid y_T = y$. 
A natural approach is to use importance sampling:
sample $(T, y_{0:T-1})$, then weight by $p(y_T = y \mid y_{T-1})$.
However, this
throws away a lot 
of work --- we make $T$ MCMC transitions but
obtain only one sample $(T, y_{0:T})$ with which to estimate the gradient.

We would like to ideally make use of all the 
MCMC transitions when constructing our estimate of $(T, y_{0:T}) \mid y_T = y$. 
For any $t \leq T$, the pair $(t, y_{0:t})$ would itself have been a valid sample 
under different randomness, and we would like to exploit this.
Suppose that we sample $T$ from some distribution $F$ and let $q(t)$ be the 
probability that $T \geq t$ under $F$. Then we can use the following scheme:
\begin{itemize}
\item Sample $T$ from $F$, then sample $y_{0:T-1}$.
\item For $t=0,\ldots,T$, weight $(t,y_{0:t-1})$ by $\frac{\epsilon (1-\epsilon)^t}{q(t)} \times p(y_t = y \mid y_{t-1})$.
\end{itemize} 
For any $q$, this yields unbiased (although unnormalized) weights (see 
Section~\ref{sec:importance} in \apx).
Typically we will choose 
$q(t) = (1-\epsilon)^t$, e.g. $F$ is a geometric distribution. If the 
$y_t$ are perfectly correlated, this will not be any more effective 
than vanilla importance sampling, but in practice this method should perform 
substantially better. Even though we obtain weights on all of
$y_{0:T}$, these weights will typically be highly correlated, so we should 
still repeat the sampling procedure multiple times 
to minimize the bias from estimating the normalization constant. The full 
procedure is given as pseudocode in Algorithm~\ref{alg:compute-gradient}.
\begin{algorithm}
\caption{Algorithm for computing an estimate of $\frac{\partial}{\partial \theta} \log p_{\theta}(y \mid x)$. This estimate is unbiased in the limit of infinitely many samples $k$, 
but will be biased for a finite number of samples due to variance in the 
estimate of the partition function.}
\label{alg:compute-gradient}
\begin{algorithmic}
\STATE {\bfseries SampleGradient}($x$, $y$, $\theta$, $\epsilon$, $k$)
\STATE \Comment $k$ is the number of samples to take
\STATE $Z \gets 0 ; g \gets 0 \quad$ \Comment $Z$ is the total importance mass of all samples, $\frac{g}{Z}$ is the gradient
\FOR{$i = 1$ {\bfseries to} $k$} 
  \STATE Sample $T \sim \Geometric(\epsilon)$
  \STATE Sample $y_0$ from $u_{\theta}(\cdot \mid x)$
  \STATE For $1 \leq t \leq T\!-\!1$: sample $y_t$ from \!$A_{\theta}(\cdot \!\mid\! y_{t-1}, x)$
  \STATE $w_0 \gets \epsilon \cdot u_{\theta}(y)$
  \STATE For $1 \leq t \leq T$: $w_t \gets \epsilon \cdot A_{\theta}(y \mid y_{t-1}, x)$
  \STATE $Z \gets Z + \sum_{t=0}^T w_t$
  \STATE $g\! \gets\! g + w_0 \frac{\partial \log u_{\theta}(y \mid x)}{\partial \theta}$ \\ 
  $\phantom{xxx\!} +\! \sum_{t=1}^T w_t \left( \frac{\partial \log u_{\theta}(y_0 \mid x)}{\partial \theta}\right.$ \\ 
  $\phantom{xxx\ \ } \left. +\! \sum_{s=1}^{t-1} \frac{\partial \log A_{\theta}(y_s \mid y_{s-1}, x)}{\partial \theta}\! +\! \frac{\partial \log A_{\theta}(y \mid y_{t-1}, x)}{\partial \theta}\right)$.
\ENDFOR
\STATE Output $\frac{g}{Z}$
\end{algorithmic}
\end{algorithm}

\subsection{Implementation}

With the theory above in place, we now describe some important implementation 
details of our learning algorithm. At a high level, we can just use 
Algorithm~\ref{alg:compute-gradient} to compute estimates of the gradient and then apply an 
online learning algorithm such as {\sc AdaGrad} \citep{duchi2011adagrad} to 
identify a good choice of $\theta$. Since the 
log-likelihood is a non-convex function 
of $\theta$, the initialization is important. We make the following 
(weak) assumption:
\begin{assumption}
\label{assumption:initialize}
The chains $u_{\theta}$ and $A_{\theta}$ are controlled by disjoint coordinates of 
$\theta$, and for any setting of $u_{\theta}$ there is a corresponding choice 
of $A_{\theta}$ that leaves $u_{\theta}$ invariant 
(i.e., $A_{\theta}u_{\theta} = u_{\theta}$).
\end{assumption}
In practice, Assumption~\ref{assumption:initialize} is easy to satisfy. For 
instance, suppose that $\phi : \sY \to \bR^d$ is a feature function, 
$\theta = [\theta_0 \ \theta_1] \in \bR^{d_0 + d}$ are the features controlling 
$u$ and $A$, 
and $u_{\theta_0}$ is made tractable by zeroing some features out:
$u_{\theta_0}(y) \propto \exp([\theta_0 \ \vec{0}_{d-d_0}]^{\top}\phi(y))$. Also 
suppose that $A_{\theta_1}$ is a Gibbs sampler that uses all the features: 
$A_{\theta_1}(y \mid y') \propto \exp(\theta_1^{\top}\phi(y_i, y_{\neg i}'))$, 
where $i$ is a randomly chosen coordinate of $y$. 
Then, we can satisfy Assumption~\ref{assumption:initialize} by setting 
$\theta_1 = [\theta_0 \ \vec{0}_{d-d_0}]$.

Under Assumption~\ref{assumption:initialize}, 
we can initialize $\theta$ by first training $u$ in isolation 
(which is a convex problem if $u_{\theta}$ parameterizes an exponential family), 
then initializing $A$ to leave $u$ invariant; this guarantees that the initial log-likelihood 
is what we would have obtained by just using $u$ by itself.
We found this to work well empirically.

As another note, Algorithm~\ref{alg:compute-gradient} 
na\"{i}vely looks like it takes $O(T^2)$ time to compute the gradient for each sample, 
due to the nested sum. However, 
most terms are of the form $w_t \frac{\partial \log A_{\theta}(y_s \mid y_{s-1}, x)}{\partial \theta}$; by grouping them for a fixed 
$s$ we can compute 
the sum in $O(T)$ time, leading to expected runtime $O\left(\frac{k}{\epsilon}\right)$ for
Algorithm~\ref{alg:compute-gradient} 
(since $\bE[T+1] = \frac{1}{\epsilon}$).

\section{Experiments}
\label{sec:experiments}
We validated our method on two challenging inference tasks. 
These tasks are difficult due to the importance of high-arity factors;
local information is insufficient to even identify high-probability regions of
the space.

\begin{figure}
\begin{center}
\raisebox{-0.5\height}{\includegraphics[width=0.9\columnwidth]{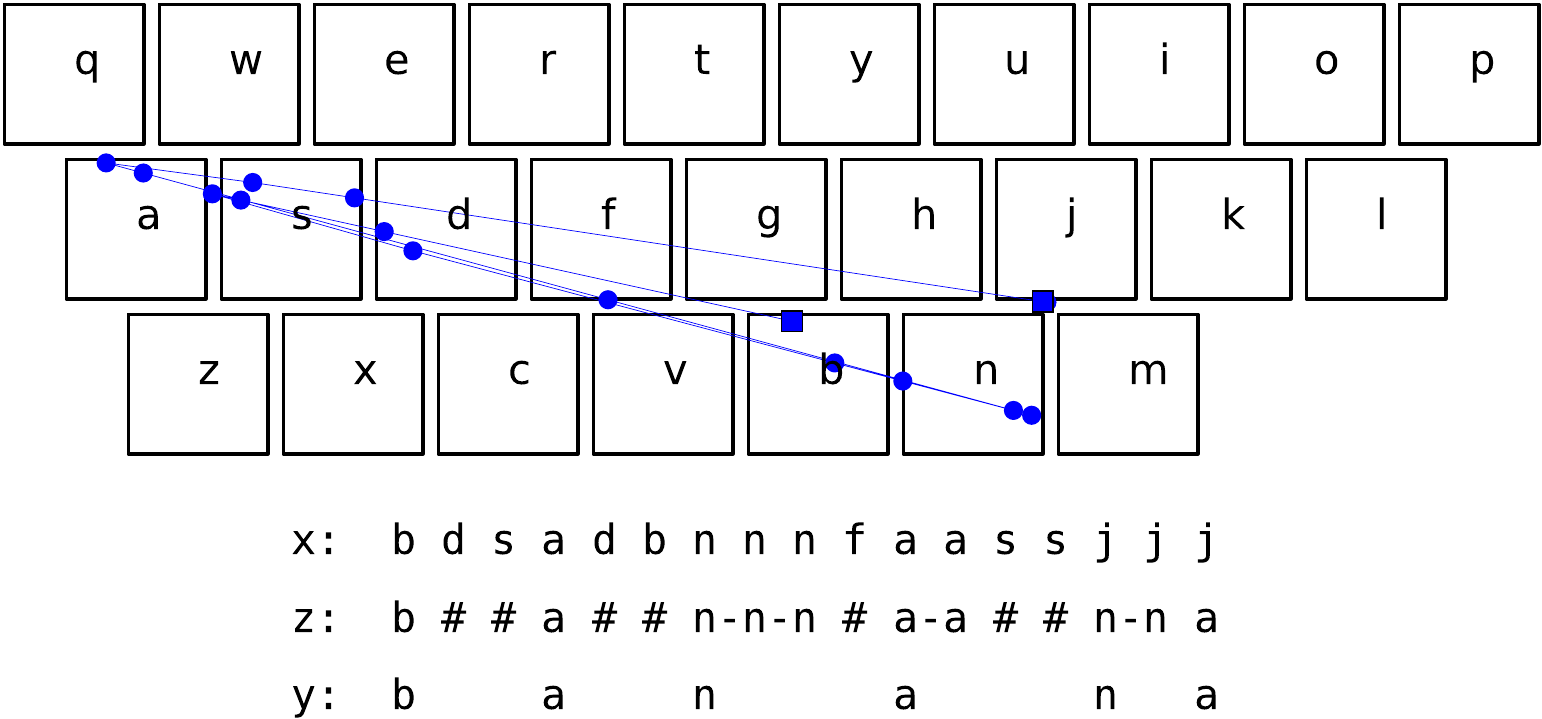}}
\end{center}
\caption{Generated sequence of keyboard gestures for the word \emph{banana}. 
The input $x$ is a sequence of characters (the recorded key presses), and the output $y$ is the 
intended word. Most characters in $x$ are incidental and do not correspond 
to any character in $y$; this is reflected by the (unobserved) alignment $z$.}
\label{fig:keyboard}
\vspace{-0.1in}
\end{figure}

\paragraph{Inferring Words from Keyboard Gestures}
We first considered the 
task of inferring words from keyboard gestures. We generated the data by sampling 
words from the New York Times corpus \citep{sandhaus2008new}.
For each word, we used a time series model to synthetically 
generate finger gestures for the word. A typical instantiation of this process 
is given in Figure~\ref{fig:keyboard}. The learning task is to discriminatively 
infer the intended word $y$ given the sequence of keys $x$ that the finger was over 
(for instance, predicting \verb=banana= from \verb=bdsadbnnnfaassjjj=). 
In our model, we posit a latent alignment $z$ 
between key presses and intended letter. Given an input $x$ of length $l$, the alignment 
$z$ also has length $l$; each $z_i$ is either `$c$' ($x_i$ starts an output letter $c$), 
`$\verb=-=c$' ($x_i$ continues an output letter $c$), or `\verb=#=' ($x_i$ is unaligned); 
see Figure~\ref{fig:keyboard} for an example. Note that $y$ is a deterministic function of $z$.

The base model $u_\theta$ consists of indicator features on $(x_i,z_i)$, $(x_i,z_{i-1},z_{i})$, 
and $(x_i,x_{i-1},z_i)$. 
The full $A_\theta$ is a Gibbs sampler in a model where we include the following features 
in addition to those above:
\begin{itemize}[itemsep=2pt,topsep=0pt,parsep=0pt,partopsep=0pt,leftmargin=12pt]
\item indicator features on $(x_i,y_i,y_{i-1})$
\item indicator of $y$ being in the dictionary, as well as log of word frequency (conditioned 
      on being in the dictionary)
\item for each $i$, indicator of $y_{1:i}$ matching a prefix of a word in the dictionary
\end{itemize}

We compared three approaches:
\begin{itemize}[itemsep=2pt,topsep=0pt,parsep=0pt,partopsep=0pt,leftmargin=12pt]
\item Our approach (\emph{Doeblin sampling})
\item Regular Gibbs sampling, initialized by setting $z_i = x_i$ for all $i$ (\emph{basic-Gibbs})
\item Gibbs sampling initialized from $u_{\theta}$ (\emph{$u_{\theta}$-Gibbs})
\end{itemize}
At test time, all three of these methods are almost identical: they all initialize from 
some distribution, then make a certain number of Gibbs samples. For basic-Gibbs and 
$u_{\theta}$-Gibbs, this is always a fixed number of steps $T$, while for Doeblin sampling 
the number of steps is a geometric distribution with mean $T$.

The main difference is in how the methods are trained. Our method is trained using the ideas 
in Section~\ref{sec:learning}; for the other two methods, we train by approximating the gradient:
\begin{align*}
\nabla \log p_{\theta}(y \mid x) &= \bE_{\hat{z} \sim p_{\theta}(z \mid x,y)}[\phi(y,\hat{z},x)] \\
 &\phantom{==} - \bE_{\hat{y},\hat{z} \sim p_{\theta}(y,z \mid x)}[\phi(\hat{y},\hat{z},x)],
\end{align*}
where $\phi(y,z,x)$ is the feature function and $p_{\theta}$ is the stationary 
distribution of $A_{\theta}$.
For the second term, we use MCMC samples from $A_{\theta}$ to approximate 
$p_{\theta}(y,z \mid x)$.
For the first term, we could take the subset of samples 
where $\hat{y} = y$, but this is problematic if no such 
samples exist. Instead, we reweight all samples with $\hat{y} \neq y$ 
by $\exp(-(D+1))$, where $D$ is the edit distance between $y$ 
and $\hat{y}$. We use the same reweighting approach for the Doeblin sampler, 
using this as the importance weight rather than using $A_{\theta}(y \mid y_{t-1})$ 
as in Algorithm~\ref{alg:compute-gradient}. 

\

To provide a fair comparison of the methods, we set $\epsilon$ in the 
Doeblin sampler to the inverse of the number of transitions $T$,
so that the expected number of transitions of all algorithms is the same.
We also devoted the first half of each chain to burn-in.

\begin{figure*}[t]
\begin{center}
\raisebox{-0.5\height}{\includegraphics[width=0.48\textwidth]{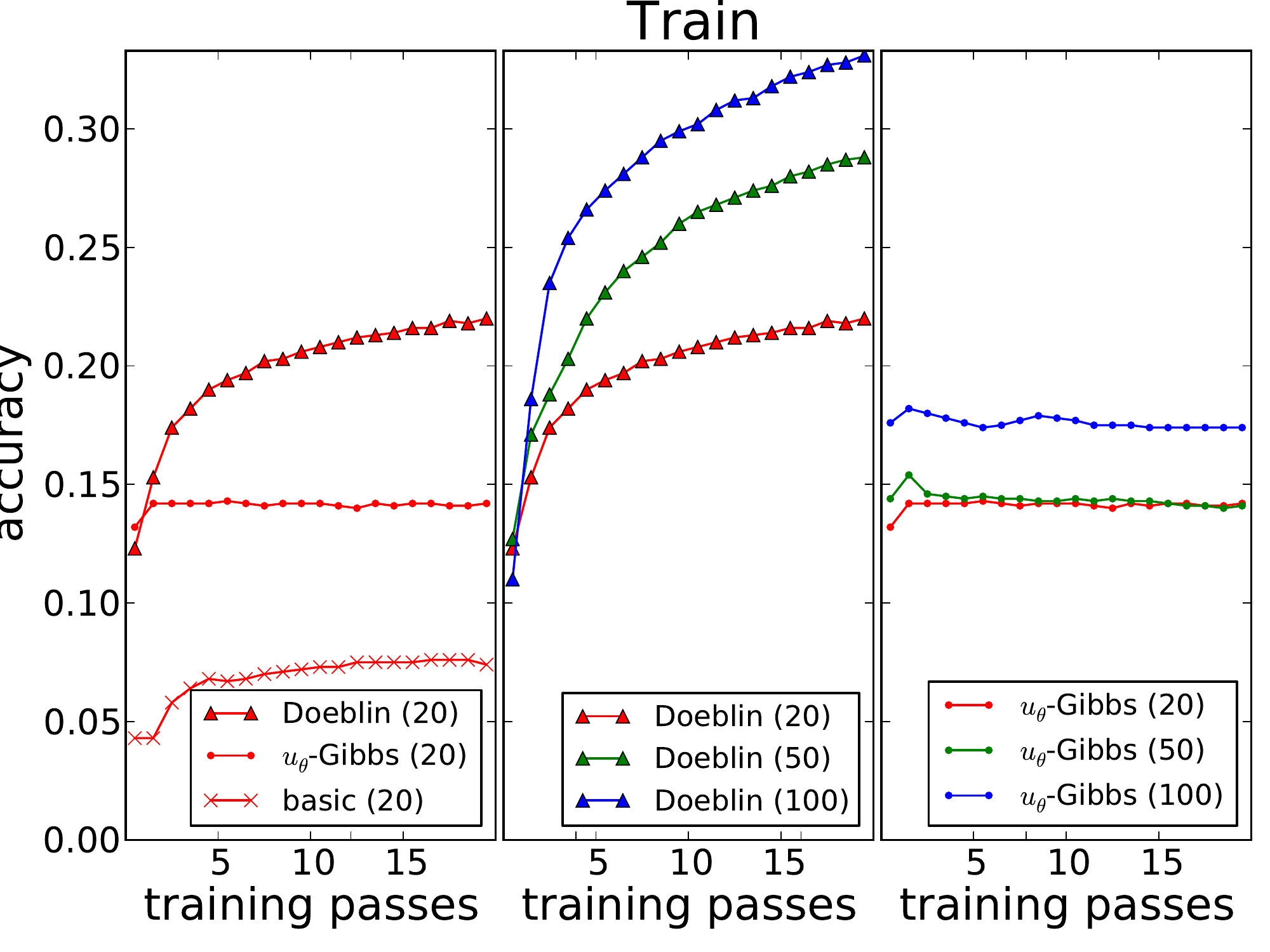}}
\hspace*{0.03in}
\raisebox{-0.5\height}{\includegraphics[width=0.48\textwidth]{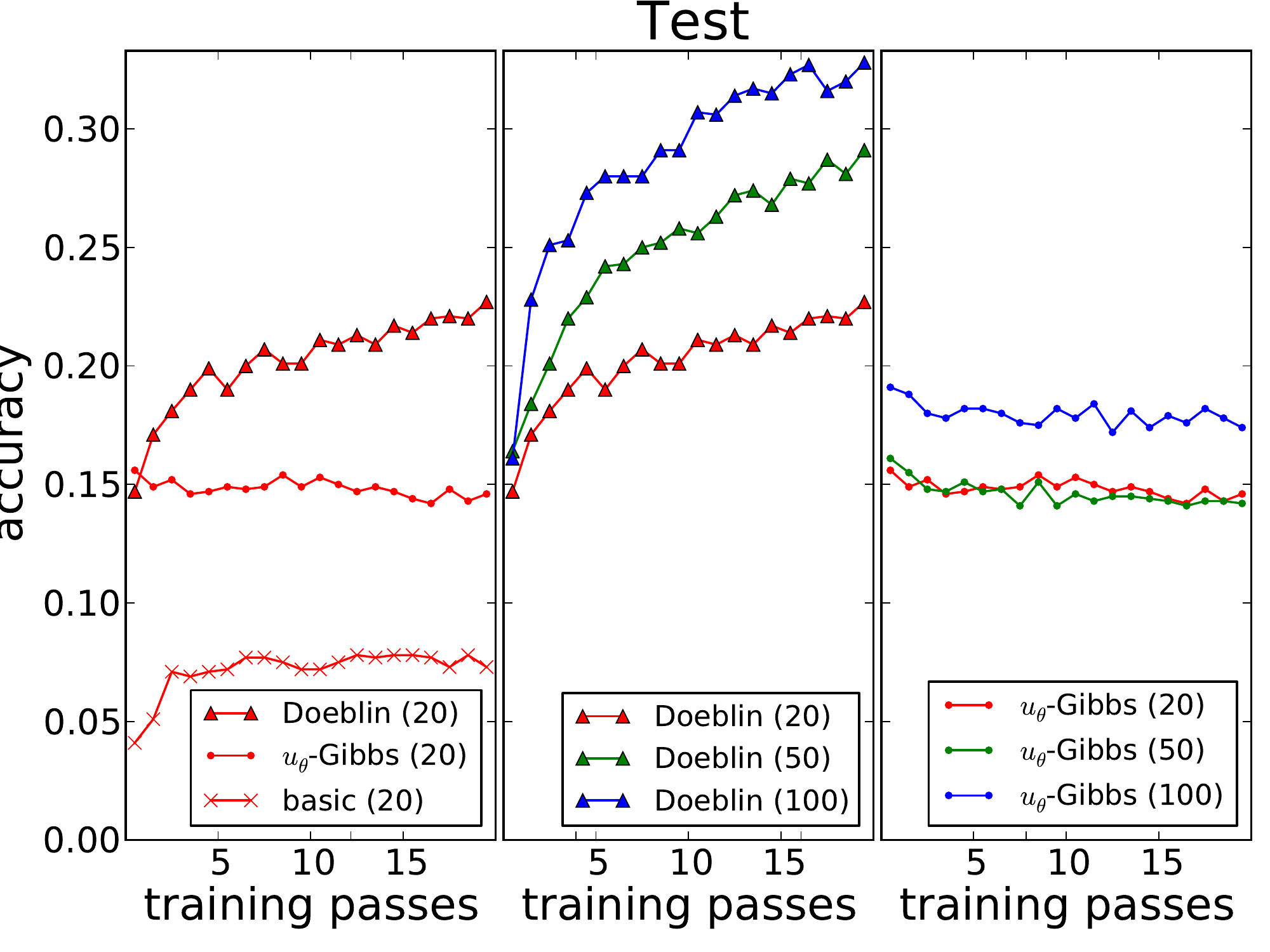}}
\end{center}
\caption{Plots of word-level (left) and character-level (right) accuracy. 
The first panel gives the performance of all $3$ methods (Doeblin chains, 
smart restarts, and simple Gibbs) for a computational budget of $20$ transitions 
per example. The second and third panels give the performance of Doeblin chains 
and smart restarts, respectively, for increasing computational budgets 
($20$, $50$, and $100$ transitions). 
}
\label{fig:gesture}
\vspace{-0.1in}
\end{figure*}

All algorithms are trained with AdaGrad \citep{duchi2011adagrad} with $16$ independent 
chains run for each example. We measure word-level accuracy by computing the fraction 
of (non-burn-in) samples whose output $y$ is correct.
\

The results are reported in Figure~\ref{fig:gesture}.
Overall, our Doeblin sampler outperforms $u_{\theta}$-Gibbs by a significant margin, 
which in turn outperforms basic-Gibbs.
Interestingly, while the accuracy of our method continues to improve with more 
training time, $u_{\theta}$-Gibbs quickly asymptotes and then slightly decreases,
even for training accuracy.

What is happening to $u_{\theta}$-Gibbs? 
Since the inference problem in this task is hard, the samples 
provide a poor gradient approximation.
As a result, optimization methods that take the approximation at face value may not converge to even a 
local optimum. This phenomenon has already been studied in other contexts, for instance 
by \citet{kulesza2007structured} and \citet{huang2012structured}.

In contrast, 
our method directly optimizes the log-likelihood 
of the data under the distribution $\tilde{\pi}_{\theta}$, so that accuracy continues to 
increase with more passes through the training data. This demonstrates that the 
MCMC samples do provide enough signal to train from, but that na\"{i}vely 
plugging them into a method designed for exact inference will fail to exploit
that signal.

\paragraph{Inferring DNF Formulas}
We next study the use of our staged Doeblin chain construction as a tool for hierarchical 
initialization. We ignore learning for now, instead treating MCMC as a stochastic search 
algorithm. Our task of interest is to infer a DNF formula $f$ from its input-output 
behavior.  This is an important subroutine in loop invariant 
synthesis, where MCMC methods have 
recently shown great promise \citep{gulwani2007program,sharmainvariant}.

Concretely, the input $x$ might look like this:
\begin{align*}
f(1,2,3) &= \mathtt{True} \\
f(1,4,4) &= \mathtt{True} \\
f(0,1,0) &= \mathtt{False} \\
f(0,2,2) &= \mathtt{True}
\end{align*}
Our task is to reconstruct $f$;
in this case, $f(x_1,x_2,x_3) = [x_1 \neq 0] \vee [x_2 = x_3]$.

More formally, we consider DNF formulae with linear inequality predicates:
$f(x) = \bigvee_{i=1}^n \bigwedge_{j=1}^m [a_{ij}^{\top}x \leq b_{ij}],$
where $a_{ij}, x \in \bZ^d$ and $b_{ij} \in \bZ$.
The formula $f$ maps input vectors to $\{\mathtt{True},\mathtt{False}\}$. 
Given a collection of example inputs and outputs, our goal is to find 
an $f$ consistent with all examples.
Our evaluation metric is the time to find such a formula.

The search space for this problem is extremely large. Even if we set 
$n = m = 3$ and restrict our search to $a_{ij} \in \{-1,0,1\}^5$, 
$b \in \{-1,0,1\}$, 
the total number of candidate formulae is still 
$\left(3^6\right)^{3 \times 3} \approx 5.8 \times 10^{25}$.

We consider three MCMC methods: no restarts ($0$-stage), uniformly 
random restarts ($1$-stage), and a staged method ($2$-stage) as in 
Section~\ref{sec:hierarchical}. 
All base chains perform Metropolis-Hastings using proposals
that edit individual atoms (e.g., $[a_{ij}^\top x \le b_{ij}]$), either by 
changing a single entry of $[a_{ij} \ b_{ij}]$ or by changing all entries of 
$[a_{ij}\ b_{ij}]$ at once.
For the staged method, we initialize uniformly at random, 
then take $\Geometric(0.04)$ transitions based on a simplified 
cost function, then take $\Geometric(0.0002)$ steps 
with the full cost (this is the staged Doeblin chain in 
Figure~\ref{fig:hierarchical}). 

The full cost function is $I(f)$, the number of examples $f$ errs on.
We stop the Markov chain when it finds a formula with $I(f) = 0$. 
The simplified cost function decomposes
over the disjuncts: for each disjunct $d(x)$, if $f(x)=\mathtt{False}$ 
while $d(x)=\mathtt{True}$, we incur a large cost (since 
in order for $f(x)$ to be false, all disjuncts comprising $f(x)$ 
must also be false). If $f(x)=\mathtt{True}$ while $d(x)=\mathtt{False}$, 
then we incur a smaller cost. If $f(x)=d(x)$ then we incur no cost.

We used all three methods as a subroutine in verifying properties of C programs; 
each such verification requires solving many instances of DNF formula inference. 
Using the staged method we are able to obtain a $30\%$ speedup 
over uniformly random restarts and a 50x improvement over no 
restarts, as shown in Table~\ref{fig:mcmc}.
\begin{table*}
\caption{Comparison of 3 different MCMC algorithms. 
0-stage uses no restarts, 1-stage uses random restarts, and 2-stage uses 
random restarts followed by a short period of MH with a 
simplified cost function. The table gives mean time and standard error (in seconds)
 taken to verify 5 different C programs, for $1000$ trials. Each verification 
requires inferring many DNF formulae as a sub-routine.}
\label{fig:mcmc}
\begin{center}

\begin{tabular}{|c|c|c|c|c|c|}
\hline 
Task & fig1 & cegar2 & nested & tacas06 & hard \\ \hline
0-stage & $2.6 \pm 1.0$ & $320 \pm 9.3$ & $120 \pm 7.0$ & $\geq 600$ & $\geq 600$ \\ \hline
1-stage & $0.074 \pm 0.001$ & $0.41 \pm 0.01$ & $2.4 \pm 0.10$ & $6.8 \pm 0.15$ & $52 \pm 1.5$ \\ \hline
2-stage & $0.055 \pm 0.005$ & $0.33 \pm 0.007$ & $2.3 \pm 0.12$ & $4.6 \pm 0.12$ & $31 \pm 0.90$ \\ \hline
\end{tabular}

\end{center}
\vspace{-0.1in}
\end{table*}

\section{Discussion}
\label{sec:discussion}

We have proposed a model family based on strong Doeblin Markov chains, which
guarantee fast mixing.
Our construction allows us to 
simultaneously leverage a simple, tractable model ($u_{\theta}$) that provides coverage together with a 
complex, accurate model ($A_{\theta}$) that provides precision.
As such, we sidestep a typical dilemma---whether to use a 
simple model with exact inference, or to deal with the consequences of
approximate inference in a more complex model.

While our approach works well in practice, there are still some 
outstanding issues. One is the non-convexity of the learning 
objective, which makes the procedure dependent on initialization. Another 
issue is that the gradients returned by Algorithm~\ref{alg:compute-gradient} 
can be large, heterogeneous, and high-variance. The adaptive nature of 
\textsc{AdaGrad} 
alleviates this somewhat, but it would still be ideal to have a sampling 
procedure that had lower variance than Algorithm~\ref{alg:compute-gradient}.

Though Gibbs sampling is the de facto method for many practitioners, 
there are also many more sophisticated approaches to MCMC \cite{green1995reversible,
earl2005parallel}. Since 
our framework is orthogonal to the particular choice of transition kernel,
it would be interesting to 
apply our method in these contexts. 

Finally, we would like to further explore the staged construction 
from Section~\ref{sec:hierarchical}. As the initial results on DNF formula
synthesis are promising,
it would be interesting to apply the 
construction to high-dimensional feature spaces as well as rich, 
multi-level hierarchies. We believe this might be a promising approach for extremely 
rich models in which a single level of re-initialization is 
insufficient to capture the complexity of the cost landscape.

\paragraph{Related work.} 
Our learning algorithm 
is reminiscent of policy gradient algorithms in reinforcement learning 
\citep{sutton00policy}, as well as Searn, which tries to learn an optimal 
search policy for structured prediction \citep{daume2009search}; see also 
\citet{shi2015learning}, who apply reinforcement learning in the context of MCMC.
Our staged construction is also 
similar in spirit to path sampling 
\citep{gelman1998simulating}, as it uses a multi-stage approach to smoothly 
transition from a very simple to a very complex distribution. 

Our staged Doeblin construction belongs to the family of 
coarse-to-fine inference methods, 
which operate on progressively more complex models
\citep{viola2004robust,
shen2004discriminative,collins2005discriminative,
charniak2006multilevel,
carreras2008tag,gu2009recognition,weiss2010sidestepping,
sapp2010cascaded,petrov2011coarse,yadollahpour2013discriminative}.

On the theoretical front, we make use of the well-developed theory of \emph{strong 
Doeblin chains}, often also referred to with the terms \emph{minorization} or 
\emph{regeneration time} \citep{doeblin1940elements,roberts1999bounds,meyn1994computable,athreya1978new}. 
The strong Doeblin property is typically used 
to study convergence of continuous-space Markov chains, but 
\citet{rosenthal1995minorization} has 
used it to analyze Gibbs sampling, and several authors 
have provided algorithms for sampling exactly from arbitrary strong Doeblin 
chains \citep{propp1996exact,corcoran1998perfect,murdoch1998exact}.
We are the first to use strong Doeblin properties to construct model families
and learn them from data.

\
\

At a high level, our idea is to identify a family of models for which an approximate inference 
algorithm is known to work well, thereby constructing a computationally tractable model family 
that is nevertheless more expressive than typical tractable families such as low-tree-width 
graphical models. We think this general program is very interesting, and could be applied to 
other inference algorithms as well, thus solidfying the link between statistical theory 
and practical reality.

\bibliographystyle{plainnat}
\bibliography{references}

\onecolumn
\appendix

\section{Proofs}

\begin{proof}[Proof of Proposition~\ref{prop:mixing}]
We will in fact show that, for all $k > 1$, $\lambda_k(\tilde{A}) = (1-\epsilon)\lambda_k(A)$, with the same eigenvector for both matrices.
In other words, while 
the stationary distribution of $\tilde{A}$ is different from $A$, all other 
eigenvectors are unchanged.

Let $w_k$ be the eigenvector of $A$ corresponding to $\lambda_k$.
First note that $1^{\top}w_k = 0$. This 
is because 
\[ 1^{\top}Aw_k = 1^{\top}w_k \]
since $A$ is stochastic, and
\[ 1^{\top}Aw_k = \lambda_k1^{\top}w_k \]
since $w_k$ is an eigenvector of $A$. Since $\lambda_k \neq 1$, this implies that $1^{\top}w_k = 0$.

Now we have
\begin{align*}
\tilde{A}w_k &= (1-\epsilon)Aw_k + \epsilon u1^{\top}w_k \\
 &= (1-\epsilon)\lambda_k w_k,
\end{align*}
which proves that $\lambda_k(\tilde{A}) = (1-\epsilon)\lambda_k(A)$. In particular, 
$\lambda_2(\tilde{A}) = (1-\epsilon)\lambda_2(A) \leq 1-\epsilon$.
The mixing 
time of $\tilde{A}$ is $\frac{1}{1-\lambda_2(\tilde{A})}$, and is therefore upper bounded by 
$\frac{1}{\epsilon}$, which completes the proof.
\end{proof}

\begin{proof}[Proof of Proposition~\ref{prop:stationary}]
We can verify algebraically that $\tilde{A}\tilde{\pi} = \tilde{\pi}$, as follows:
\begin{align*}
\tilde{A}\tilde{\pi} &= (1-\epsilon)A\tilde{\pi} + \epsilon u1^{\top}\tilde{\pi} \\
 &= \epsilon(1-\epsilon)A(I-(1-\epsilon)A)^{-1}u + \epsilon u \\
 &= \epsilon \left[(1-\epsilon)A (I-(1-\epsilon)A)^{-1} + I\right] u \\
 &= \epsilon \left[(1-\epsilon)A + (I-(1-\epsilon)A) \right] (I-(1-\epsilon)A)^{-1} u \\
 &= \epsilon (I - (1-\epsilon)A)^{-1} u \\
 &= \tilde{\pi},
\end{align*}
so that $\tilde{\pi}$ is indeed the stationary distribution of $\tilde{A}$. 
\end{proof}

\begin{proof}[Proof of Proposition~\ref{prop:mahalanobis}]
From the characterization of $\tilde{\pi}$ in Proposition~\ref{prop:stationary}, 
we know  that $\tilde{\pi}$ is equal to
\[ \epsilon \sum_{j=0}^{\infty} (1-\epsilon)^j A^j u. \]
The rest of the proof consists of determining some useful properties of 
$d_{\pi}(\pi')$. The most important (and the motivation for defining 
$d_{\pi}$ in the first place) is given in the following lemma, which we prove separately:
\begin{lemma}
\label{lem:contract}
If $\pi$ is the stationary distribution of $A$ and $A$ satisfies detailed balance, then
$d_{\pi}(A\pi') \leq \lambda_2(A) d_\pi(\pi')$. 
\end{lemma}
The other important property of $d_{\pi}(\pi')$ is convexity: 
$d_{\pi}(w \pi' + (1-w)\pi'') \leq w d_{\pi}(\pi') + (1-w)d_{\pi}(\pi'')$, 
which follows directly from the characterization of $d_{\pi}(\pi')$ as a 
Mahalanobis distance.

Putting these two properties together, we have
\begin{align*}
d_{\pi}(\tilde{\pi}) &\leq \epsilon \sum_{j=0}^{\infty} (1-\epsilon)^j d_{\pi}(A^ju) \\
 &\leq \epsilon \sum_{j=0}^{\infty} (1-\epsilon)^j \lambda_2(A)^j d_{\pi}(u) \\
 &\leq \frac{\epsilon}{1-(1-\epsilon)\lambda_2(A)}d_{\pi}(u) \\
 &\leq \frac{\epsilon}{1-\lambda_2(A)}d_{\pi}(u),
\end{align*}
which completes the proof.
\end{proof}
\begin{proof}[Proof of Lemma~\ref{lem:contract}]
Recall we want to show that $d_{\pi}(A\pi') \leq \lambda_2(A) d_\pi(\pi')$. 
To see this, first define $S = \diag(\pi)^{-1/2}A\diag(\pi)^{1/2}$, which is symmetric 
by the detailed balance condition, and satisfies $\lambda_k(S) = \lambda_k(A)$ by 
similarity. Furthermore, the top eigenvector of $S$ is $\bi^{\top}\diag(\pi)^{1/2}$. 
Putting these together, we have
\begin{align*}
d_{\pi}(A\pi') &= \|\diag(\pi)^{-1/2}(\pi-A\pi')\|_2 \\
 &= \|\diag(\pi)^{-1/2}A(\pi-\pi')\|_2 \\
 &= \|S\diag(\pi)^{-1/2}(\pi-\pi')\|_2 \\
 &\leq \lambda_2(S) \|\diag(\pi)^{-1/2}(\pi-\pi')\|_2 \\
 &= \lambda_2(S) d_{\pi}(\pi') \\
 &= \lambda_2(A) d_{\pi}(\pi').
\end{align*}
The inequality step $\|S\diag(\pi)^{-1/2}(\pi-\pi')\|_2 \leq \lambda_2(S) \|\diag(\pi)^{-1/2}(\pi-\pi')\|_2$ follows because $\diag(\pi)^{-1/2}(\pi-\pi')$ is orthogonal to the top eigenvector 
$\bi^{\top}\diag(\pi)^{1/2}$ of $S$.
\end{proof}
\begin{proof}[Proof of Proposition~\ref{prop:monotonic}]
For any value of $\epsilon$, the stationary distribution $\tilde{\pi}_{\epsilon}$ of 
$(1-\epsilon)A + \epsilon u \bi^{\top}$ can be obtained by applying $A$ a 
$\Geometric(\epsilon)$-distributed number of times to $u$ (by Proposition~\ref{prop:stationary}).
For $\epsilon_2 < \epsilon_1$, it therefore suffices to construct a random variable 
$F \geq 0$ such that 
if $s \sim F$ and $t \sim \Geometric(\epsilon_1)$ then $s+t \sim \Geometric(\epsilon_2)$; if we 
can do this, then we can let $B$ be the matrix that applies $A$ an $F$-distributed number of 
times, and we would have $B\tilde{\pi}_{\epsilon_1} = \tilde{\pi}_{\epsilon_2}$; but $B$ would 
clearly have stationary distribution $\pi$, and so Lemma~\ref{lem:klmc} would give 
$\KL{\pi}{\tilde{\pi}_{\epsilon_2}} \leq \KL{\pi}{\tilde{\pi}_{\epsilon_1}}$ and 
$\KL{\tilde{\pi}_{\epsilon_2}}{\pi} \leq \KL{\tilde{\pi}_{\epsilon_1}}{\pi}$, which is the 
desired result.

To construct the desired $F$, we use the fact that addition of random variables corresponds 
to convolution of the probability mass functions, and furthermore represent the probability 
mass functions as formal power series; in particular, we let
\[ f(x) = \sum_{n=0}^{\infty} \bP[t = n \mid t \sim F]x^n, \]
and similarly
\begin{align*}
g_{\epsilon}(x) &= \sum_{n=0}^{\infty} \bP[t = n \mid t \sim \Geometric(\epsilon)]x^n \\
 &= \sum_{n=0}^{\infty} \epsilon(1-\epsilon)^n x^n \\
 &= \frac{\epsilon}{1-(1-\epsilon)x}.
\end{align*}
We want $f(x)g_{\epsilon_1}(x)$ to equal $g_{\epsilon_2}(x)$, so we take
\begin{align*}
f(x) &= \frac{g_{\epsilon_2}(x)}{g_{\epsilon_1}(x)} \\
 &= \frac{\epsilon_2}{\epsilon_1} \frac{1-(1-\epsilon_1)x}{1-(1-\epsilon_2)x} \\
 &= \frac{\epsilon_2}{\epsilon_1} \left(1 + \sum_{n=1}^{\infty} \left[(1-\epsilon_2)^n - (1-\epsilon_2)^{n-1}(1-\epsilon_1)\right]x^n\right) \\
 &= \frac{\epsilon_2}{\epsilon_1} \left(1 + \sum_{n=1}^{\infty} (1-\epsilon_2)^{n-1}(\epsilon_1-\epsilon_2) x^n \right).
\end{align*}
From this we see that the random variable $F$ with probability function 
\[ \bP[t = n \mid t \sim F] = \left\{ \begin{array}{ccl} \frac{\epsilon_2}{\epsilon_1} & : & n = 0 \\
 \frac{\epsilon_2}{\epsilon_1}(1-\epsilon_2)^{n-1}(\epsilon_1-\epsilon_2) & : & n > 0 \end{array} \right. \]
satisfies the required property that $F + \Geometric(\epsilon_1) = \Geometric(\epsilon_2)$. Note 
that the condition $\epsilon_2 < \epsilon_1$ is necessary here so that all of the probability 
masses are positive in the above expression.

We can also prove the result purely algebraically. Motivated by the above construction, we 
define 
\begin{align*}
B &= \epsilon_2(I-(1-\epsilon_2)A)^{-1}\left[\epsilon_1(I-(1-\epsilon_1)A)^{-1}\right]^{-1} \\
  &= \frac{\epsilon_2}{\epsilon_1}\left[I + (\epsilon_1 - \epsilon_2)A(I-(1-\epsilon_2)A)^{-1}\right].
\end{align*}
By construction we have $B\tilde{\pi}_{\epsilon_1} = \tilde{\pi}_{\epsilon_2}$, 
but Taylor expanding the second expression for $B$ shows that 
we can write it as a (infinite) convex combination of non-negative powers of 
$A$, and hence that $B$ has stationary distribution $\pi$. This again yields the 
desired result by Lemma~\ref{lem:klmc}.
\end{proof}
\begin{proof}[Proof of Theorem~\ref{thm:hierarchical}]
We use an equivalent characterization of the strong Doeblin parameter as the 
quantity $\Gamma(A) = \sum_{y} \inf_{y'} A(y \mid y')$. 
In the context of the Markov 
chain $M$, this yields
\begin{align*}
\Gamma(M^b) &= \sum_{(z_b,y_b)} \inf_{(z_0, y_0)} \bP[z_b,y_b \mid z_0, y_0] \\
 &= \sum_{(z_b,y_b)} \inf_{(z_0, y_0)} \sum_{\tau=a}^b \bP[\tau_a = \tau \mid y_0] \times \bP[z_b, y_b \mid \tau_a = \tau] \\
 &\geq \sum_{(z_b, y_b)} \sum_{\tau=a}^b \inf_{y_0} \bP[\tau_a = \tau \mid y_0] \times \bP[z_b, y_b \mid \tau_a = \tau] \\
 &= \sum_{\tau=a}^b \inf_{y_0} \bP[\tau_a = \tau \mid y_0] \sum_{(z_b, y_b)} \bP[z_b, y_b \mid \tau_a = \tau] \\
 &= \sum_{\tau = a}^b \inf_{y_0} \bP[\tau_a = \tau \mid y_0] \\
 &= \gamma_{a,b}.
\end{align*}
Finally, by Proposition~\ref{prop:mixing}, the spectral gap of $M^b$ is at least 
$\gamma_{a,b}$, hence the spectral gap of $M$ is at least 
$\frac{1}{b}\gamma_{a,b}$, which proves the theorem.
\end{proof}

\begin{proof}[Proof of Corollary~\ref{cor:hierarchical}]
Note that the time 
to transition from $i$ to $i+1$ is $\Geometric(\delta_i)$-distributed. 
Suppose we start from an arbitrary $j \in \{0,\ldots,k-1\}$. Then the time $t'$ to 
get to $k-1$ is distributed as $\sum_{i=j}^{k-2} \Geometric(\delta_i)$. 
$t'$ has mean $\sum_{i=j}^{k-2} \frac{1}{\delta_i} \leq \frac{1}{\delta_{k-1}}$,
 and variance $\sum_{i=j}^{k-2} \frac{1-\delta_i}{\delta_i^2} \leq \frac{1}{2\delta_{k-1}^2}$. In particular, with probability 
$\frac{1}{2}$, $t'$ lies between $0$ and 
$\left\lfloor\frac{2}{\delta_{k-1}}\right\rfloor$. Now, consider the time $t''$ to 
get from $k-1$ to $0$; this is $\Geometric(\delta_{k-1})$-distributed, and 
conditioned on $t''$ being at least $\frac{2}{\delta_{k-1}}$, 
$t'' + t' - \frac{2}{\delta_{k-1}}$ is also 
$\Geometric(\delta_{k-1})$-distributed. But $t''$ is at least 
$\lfloor\frac{2}{\delta_{k-1}}\rfloor$ with probability at least
$(1-\delta_{k-1})^{2/\delta{k-1}} \geq \frac{1}{16}$ 
(since $\delta_{k-1} \leq \frac{1}{2}$). Hence independently of 
$j$, $t'' + t'$ is, with probability at least $\frac{1}{16}$, distributed 
according to $\lfloor\frac{2}{\delta_{k-1}}\rfloor + \Geometric(\delta_{k-1})$. But 
$t'' + t' = \tau$ by construction, and so we need only compute the probability 
that $\tau \leq \lceil\frac{3}{\delta_{k-1}}\rfloor$; but this is just the 
probability that the geometric distribution is less than 
$\frac{1}{\delta_{k-1}}$, which is 
$1-(1-\delta_{k-1})^{1/\delta_{k-1}} \geq 1-e^{-1}$. 
Therefore, $\gamma(t_0, t) \geq \frac{1-e^{-1}}{16} \geq \frac{1}{26}$; expanding 
the definitions of $t_0$ and $t$, we have that
$\gamma_{\lfloor 2/\delta_{k-1}\rfloor,\lceil 3/\delta_{k-1}\rceil} \geq \frac{1}{26}$. Applying Theorem~\ref{thm:hierarchical} then implies that the spectral gap of 
$M$ is at least $\frac{\delta_{k-1}}{78}$, as was to be shown.
\end{proof}
\begin{proof}[Proof of Lemma~\ref{lem:gradient}]
The key idea is to use the identity $\frac{\partial f}{\partial x} = f(x) \frac{\partial \log f(x)}{\partial x}$ in two places. We have
\begin{align*}
p_\theta(z \in S) \frac{\partial \log p_\theta(z \in S)}{\partial \theta} &= \frac{\partial}{\partial \theta} p_\theta(z \in S) \\
 &= \int_{S} \frac{\partial p_\theta(z)}{\partial \theta} dz \\
 &= \int_{S} p_\theta(z) \frac{\partial \log p_\theta(z)}{\partial \theta} dz \\
 &= p_\theta(z \in S) \bE_z\left[ \frac{\partial \log p_\theta(z)}{\partial \theta} \middle| z \in S\right],
\end{align*}
which completes the lemma.
\end{proof}

\section{Correctness of Importance Sampling Algorithm}
\label{sec:importance}
In Section~\ref{sec:sampling} of the main text, we had a distribution $u$ over $\sY$ and a 
Markov chain $A(y_t \mid y_{t-1})$ on the same space. We then built a distribution over 
$\sY^* \eqdef \bigcup_{T \geq 0} \{T\} \times \sY^T$ by sampling $T \sim \Geometric(\epsilon)$, 
$y_0 \sim u$, and $y_t \mid y_{t-1} \sim A$ for $y=1,\ldots,T$ (we use $p_T(y_{0:T})$ to 
represent the distribution over $y_{0:T}$ given $T$).

For a given $y$, 
we were interested in constructing an importance sampler for $(T, y_{0:T}) \mid y_T = y$.
The following Lemma establishes the correctness of the importance sampler that was presented. 
We assume we are interested in computing the expectation of some function $g : \sY^* \to \bR$ 
and show that the given importance weights correctly estimate $\bE[g]$.
\begin{lemma}
\label{lem:importance}
For a distribution $F$, suppose that we sample $T \sim F$ and then sample 
$y_{0:T-1} \sim p_{T-1}$. 
Let $w_t = \frac{\epsilon(1-\epsilon)^t}{\bP[T \geq t \mid T \sim F]}A(y \mid y_{t-1})$. 
Consider the random variable 
\[ \hat{g} \eqdef \sum_{t=0}^T w_t g(t, y_{0:t-1}, y). \]
Then 
\[ \bE_{T \sim F, y_{0:T-1} \sim p_{T-1}}[\hat{g}] = \bE_{T \sim \Geometric(\epsilon), y_{0:T} \sim p_T}\left[g(T, y_{0:T})\bI[y_T = y]\right]. \]
\end{lemma}
\begin{proof}
We have
\begin{align*}
\bE_{T \sim F, y_{0:T-1} \sim p}[\hat{g}] &= \bE_{T \sim F,y_{0:T-1}\sim p_{T-1}}\left[\sum_{t=0}^T w_t g(t, y_{0:t-1}, y)\right] \\
 &= \sum_{t=0}^{\infty} \bP[T \geq t \mid T \sim F] \bE_{y_{0:t-1}\sim p_{t-1}}\left[ w_t g(t, y_{0:t-1}, y)\right] \\
 &= \sum_{t=0}^{\infty} \epsilon(1-\epsilon)^t \bE_{y_{0:t-1}\sim p_{t-1}}\left[A(y \mid y_{t-1}) g(t, y_{0:t-1}, y)\right] \\
 &= \sum_{t=0}^{\infty} \epsilon(1-\epsilon)^t \bE_{y_{0:t} \sim p_t}\left[g(t, y_{0:t})\bI[y_t = y]\right] \\
 &= \bE_{T \sim \Geometric(\epsilon), y_{0:T} \sim p_T}\left[g(T, y_{0:T})\bI[y_T=y]\right],
\end{align*}
as was to be shown.
\end{proof}

\end{document}